\documentclass[11pt]{article}
\usepackage[margin=1in]{geometry}
\usepackage[T1]{fontenc}
\usepackage{lmodern}
\usepackage{microtype}
\usepackage{amsmath,amssymb}
\usepackage{booktabs}
\usepackage{multirow}
\usepackage{graphicx}
\usepackage{xcolor}
\usepackage{adjustbox}
\usepackage{enumitem}
\usepackage{fvextra}
\DefineVerbatimEnvironment{PromptBlock}{Verbatim}{fontsize=\scriptsize,breaklines=true,breakanywhere=true,breaksymbolleft={},breaksymbolright={}}
\usepackage[numbers,sort&compress]{natbib}
\usepackage{hyperref}
\usepackage{cleveref}
\usepackage{algorithm}
\usepackage{algpseudocode}
\hypersetup{colorlinks=true,linkcolor=blue!50!black,citecolor=blue!50!black,urlcolor=blue!50!black}
\setlist{nosep,leftmargin=*}

\newcommand{\method}{\textsc{RSI-Opt}}
\newcommand{\opal}{\textsc{OPAL}}
\newcommand{\baseline}{single-agent baseline}
\newcommand{\Baseline}{Single-agent}

\newcommand{\auditFlaggedInner}{1}
\newcommand{\auditFlaggedMeta}{12}
\newcommand{\auditInnerCalls}{18,895}
\newcommand{\auditInnerTranscripts}{528}
\newcommand{\auditMetaCalls}{949}
\newcommand{\auditMetaTranscripts}{20}
\newcommand{\bestGen}{4}
\newcommand{\bestPolicy}{p016}
\newcommand{\bestScore}{+14.63}

\newcommand{\budFirstOpalHeldML}{+6.35}

\newcommand{\budFirstRsiHeldML}{+6.82}

\newcommand{\budFirstRsiTrain}{+14.56}

\newcommand{\budFrozenLibHeldML}{-21.80}

\newcommand{\budOnedecOpalHeldML}{+6.12}

\newcommand{\budOnedecRsiHeldML}{+5.05}

\newcommand{\budOnedecRsiTrain}{+13.82}

\newcommand{\budOpalHeldML}{+8.75}

\newcommand{\budRsiHeldML}{+8.26}

\newcommand{\budRsiTrain}{+14.37}

\newcommand{\decisionsPerRun}{20}
\newcommand{\dzGptFiveHeldML}{+5.39}
\newcommand{\dzGptFiveHeldPDE}{+1.97}
\newcommand{\dzGptFiveTrain}{+9.69}
\newcommand{\dzGptSixHeldML}{+6.92}
\newcommand{\dzGptSixHeldPDE}{-2.50}
\newcommand{\dzGptSixTrain}{+4.15}
\newcommand{\dzOpusHeldML}{-0.49}
\newcommand{\dzOpusHeldPDE}{+0.68}
\newcommand{\dzOpusTrain}{+3.80}

\newcommand{\genZeroScore}{+10.30}
\newcommand{\nExcluded}{2}

\newcommand{\nGenerations}{5}

\newcommand{\nPolicies}{21}
\newcommand{\scoreFourPNineteen}{+14.42}
\newcommand{\scoreFourPSixteen}{+14.55}
\newcommand{\scoreFourPTwenty}{+14.19}

\newcommand{\trajGenOneHeldML}{+9.75}

\newcommand{\trajGenOneTrain}{+11.49}
\newcommand{\trajGenThreeHeldML}{+7.51}

\newcommand{\trajGenThreeTrain}{+13.91}
\newcommand{\trajOpalHeldML}{+8.75}

\newcommand{\trajOpalTrain}{+10.57}

\newcommand{\zOpalGptFiveHeld}{+0.42}
\newcommand{\zOpalGptFiveHeldML}{+1.19}
\newcommand{\zOpalGptFiveHeldPDE}{-0.34}
\newcommand{\zOpalGptFiveTrain}{+4.43}
\newcommand{\zOpalGptSixHeld}{+3.32}
\newcommand{\zOpalGptSixHeldML}{-0.01}
\newcommand{\zOpalGptSixHeldPDE}{+6.64}
\newcommand{\zOpalGptSixTrain}{+10.09}
\newcommand{\zOpalOpusHeld}{+6.63}
\newcommand{\zOpalOpusHeldML}{+8.75}
\newcommand{\zOpalOpusHeldPDE}{+4.50}
\newcommand{\zOpalOpusTrain}{+10.57}
\newcommand{\zRsiGptFiveHeld}{+4.06}
\newcommand{\zRsiGptFiveHeldML}{+6.58}
\newcommand{\zRsiGptFiveHeldPDE}{+1.55}
\newcommand{\zRsiGptFiveTrain}{+14.12}
\newcommand{\zRsiGptSixHeld}{+5.52}
\newcommand{\zRsiGptSixHeldML}{+6.90}
\newcommand{\zRsiGptSixHeldPDE}{+4.14}
\newcommand{\zRsiGptSixTrain}{+14.24}
\newcommand{\zRsiOpusHeld}{+6.72}
\newcommand{\zRsiOpusHeldML}{+8.26}
\newcommand{\zRsiOpusHeldPDE}{+5.18}
\newcommand{\zRsiOpusTrain}{+14.37}
\newcommand{\zTunedHeld}{+1.06}
\newcommand{\zTunedHeldML}{+0.49}
\newcommand{\zTunedHeldPDE}{+1.63}
\newcommand{\zTunedTrain}{+3.30}

\providecommand{\zTunedTrain}{}
\renewcommand{\zTunedTrain}{+3.46}
\providecommand{\zTunedHeldML}{}
\renewcommand{\zTunedHeldML}{+0.49}
\providecommand{\zTunedHeldPDE}{}
\renewcommand{\zTunedHeldPDE}{+1.11}
\providecommand{\zTunedHeld}{}
\renewcommand{\zTunedHeld}{+0.80}
\providecommand{\zOpalOpusTrain}{}
\renewcommand{\zOpalOpusTrain}{+10.48}
\providecommand{\zOpalOpusHeldML}{}
\renewcommand{\zOpalOpusHeldML}{+8.71}
\providecommand{\zOpalOpusHeldPDE}{}
\renewcommand{\zOpalOpusHeldPDE}{+1.72}
\providecommand{\zOpalOpusHeld}{}
\renewcommand{\zOpalOpusHeld}{+5.22}
\providecommand{\zRsiOpusTrain}{}
\renewcommand{\zRsiOpusTrain}{+14.33}
\providecommand{\zRsiOpusHeldML}{}
\renewcommand{\zRsiOpusHeldML}{+8.23}
\providecommand{\zRsiOpusHeldPDE}{}
\renewcommand{\zRsiOpusHeldPDE}{+1.51}
\providecommand{\zRsiOpusHeld}{}
\renewcommand{\zRsiOpusHeld}{+4.87}
\providecommand{\zOpalGptFiveTrain}{}
\renewcommand{\zOpalGptFiveTrain}{+4.33}
\providecommand{\zOpalGptFiveHeldML}{}
\renewcommand{\zOpalGptFiveHeldML}{+1.18}
\providecommand{\zOpalGptFiveHeldPDE}{}
\renewcommand{\zOpalGptFiveHeldPDE}{-1.90}
\providecommand{\zOpalGptFiveHeld}{}
\renewcommand{\zOpalGptFiveHeld}{-0.36}
\providecommand{\zRsiGptFiveTrain}{}
\renewcommand{\zRsiGptFiveTrain}{+14.10}
\providecommand{\zRsiGptFiveHeldML}{}
\renewcommand{\zRsiGptFiveHeldML}{+6.54}
\providecommand{\zRsiGptFiveHeldPDE}{}
\renewcommand{\zRsiGptFiveHeldPDE}{-0.32}
\providecommand{\zRsiGptFiveHeld}{}
\renewcommand{\zRsiGptFiveHeld}{+3.11}
\providecommand{\zOpalGptSixTrain}{}
\renewcommand{\zOpalGptSixTrain}{+9.90}
\providecommand{\zOpalGptSixHeldML}{}
\renewcommand{\zOpalGptSixHeldML}{-0.01}
\providecommand{\zOpalGptSixHeldPDE}{}
\renewcommand{\zOpalGptSixHeldPDE}{+1.61}
\providecommand{\zOpalGptSixHeld}{}
\renewcommand{\zOpalGptSixHeld}{+0.80}
\providecommand{\zRsiGptSixTrain}{}
\renewcommand{\zRsiGptSixTrain}{+14.21}
\providecommand{\zRsiGptSixHeldML}{}
\renewcommand{\zRsiGptSixHeldML}{+6.87}
\providecommand{\zRsiGptSixHeldPDE}{}
\renewcommand{\zRsiGptSixHeldPDE}{+1.74}
\providecommand{\zRsiGptSixHeld}{}
\renewcommand{\zRsiGptSixHeld}{+4.30}
\providecommand{\dzOpusTrain}{}
\renewcommand{\dzOpusTrain}{+3.85}
\providecommand{\dzOpusHeldML}{}
\renewcommand{\dzOpusHeldML}{-0.48}
\providecommand{\dzOpusHeldPDE}{}
\renewcommand{\dzOpusHeldPDE}{-0.21}

\providecommand{\dzGptFiveTrain}{}
\renewcommand{\dzGptFiveTrain}{+9.77}
\providecommand{\dzGptFiveHeldML}{}
\renewcommand{\dzGptFiveHeldML}{+5.36}
\providecommand{\dzGptFiveHeldPDE}{}
\renewcommand{\dzGptFiveHeldPDE}{+1.59}

\providecommand{\dzGptSixTrain}{}
\renewcommand{\dzGptSixTrain}{+4.31}
\providecommand{\dzGptSixHeldML}{}
\renewcommand{\dzGptSixHeldML}{+6.88}
\providecommand{\dzGptSixHeldPDE}{}
\renewcommand{\dzGptSixHeldPDE}{+0.13}

\providecommand{\budOpalHeldML}{}
\renewcommand{\budOpalHeldML}{+8.71}

\providecommand{\budRsiTrain}{}
\renewcommand{\budRsiTrain}{+14.33}
\providecommand{\budRsiHeldML}{}
\renewcommand{\budRsiHeldML}{+8.23}

\providecommand{\trajOpalTrain}{}
\renewcommand{\trajOpalTrain}{+10.48}
\providecommand{\trajOpalHeldML}{}
\renewcommand{\trajOpalHeldML}{+8.71}

\providecommand{\budFirstRsiHeldML}{}
\renewcommand{\budFirstRsiHeldML}{+6.82}
\providecommand{\budOnedecRsiHeldML}{}
\renewcommand{\budOnedecRsiHeldML}{+5.05}
\providecommand{\budFirstOpalHeldML}{}
\renewcommand{\budFirstOpalHeldML}{+6.35}
\providecommand{\budOnedecOpalHeldML}{}
\renewcommand{\budOnedecOpalHeldML}{+6.12}

\title{RSI-Opt: Recursive Self-Improvement Optimizer Agents For Model Training}
\author{Binghang Lu$^{1,*}$, Yige Sun$^{2}$, Guang Lin$^{1,*}$\\[0.5em]
$^{1}$Purdue University\\
$^{2}$Independent Researcher\\[0.5em]
$^{*}$Corresponding authors}
\date{}
\begin{document}
\maketitle

\begin{abstract}
Conventional neural-network training typically follows a predefined optimization recipe, motivating a LLM agent that monitors training dynamics and adjusts optimizer settings as training progresses, but a fixed agent policy
does not accumulate experience across runs. We introduce \method{}, a plug-and-play Recursive Self-Improvement Optimizer Agents which evolves the knowledge that guides this
process: a prompt, a playbook of procedures, a library of tested optimizer code, and a decision cadence. A meta-agent
rewrites this policy using evidence from earlier runs, and an archive retains evaluated policies for further
evolution. Across \nGenerations{} generations on three training tasks, the meta-agents develop reusable procedures and code
and conduct their own CPU proxy experiments. RSI-Opt substantially improves optimizer-agent performance, with representative raw-space summary $z$-scores increasing from $+10.48$ to $+14.33$ on the evolution suite with Opus~5.5 and from $-0.36$ to $+3.11$ on held-out tasks with GPT-5.5 through cross-model policy transfer. On the Burgers PDE, it reduces mean relative $L_2$ error by approximately $2.7$-fold with Opus~5.5 and $8.8$-fold with GPT-6.1 over the fixed-prompt agent baseline, although PDE transfer varies across LLMs and tasks. Controlled comparisons show that the evolved policy combines reusable optimization knowledge with adaptation to each task: initial decisions capture most observed gains on seen tasks, while continued adaptation improves outcomes on unseen tasks. These findings demonstrate how procedures and executable tools accumulated across training runs can improve optimizer-agent performance and support transfer to scientific machine learning.

\end{abstract}

\section{Introduction}
\label{sec:intro}

\begin{figure}[t]
\centering
\includegraphics[width=\linewidth]{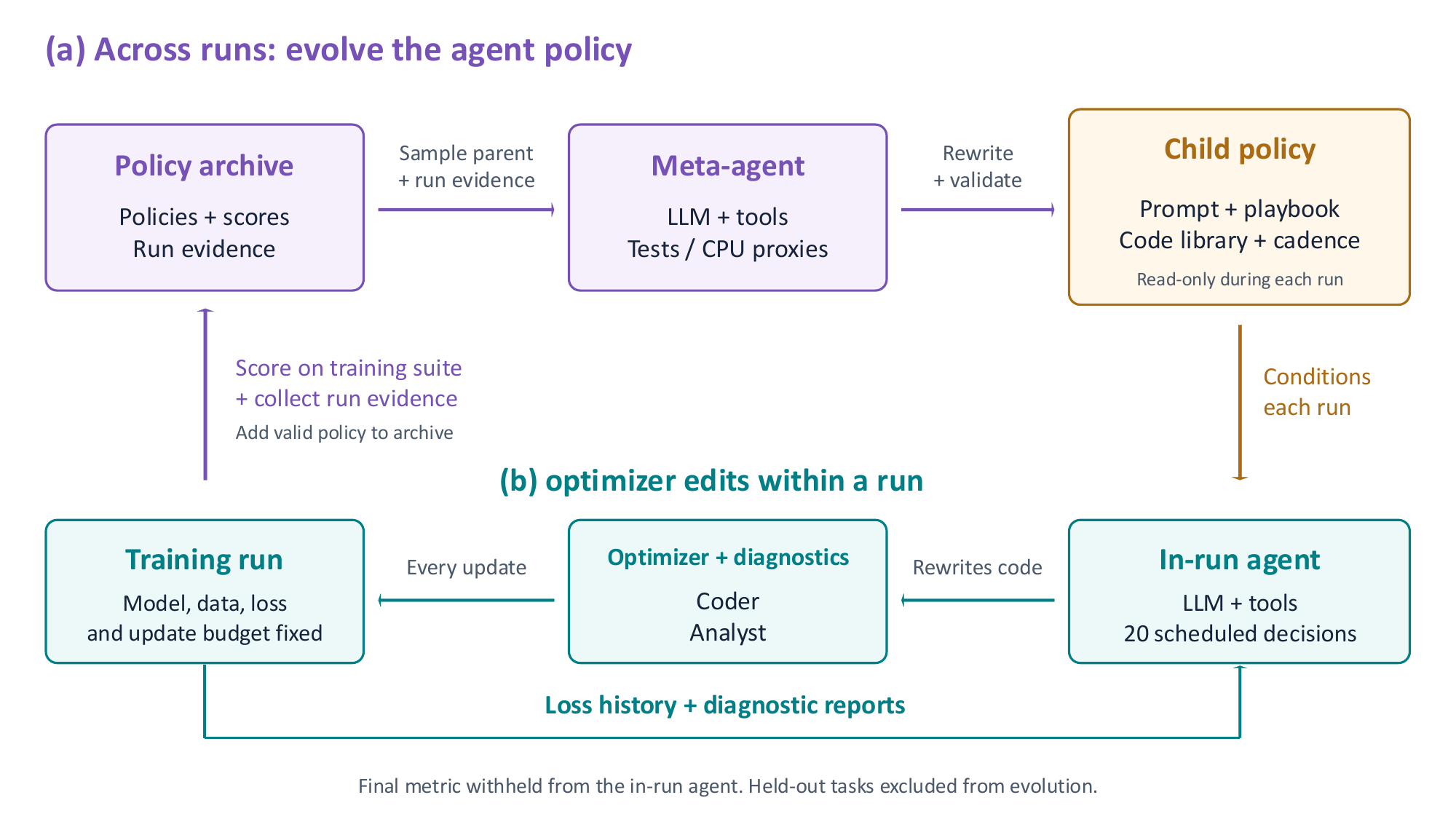}
\caption{\method{} evolves reusable knowledge for an optimizer agent. (a) Across runs, a meta-agent samples a
parent policy and run evidence, rewrites and validates a child, and evaluates it on the training suite. Every valid
evaluated child enters the archive with its score and run evidence. (b) Within a run, an agent implemented reads losses and diagnostics at scheduled decisions and rewrites the optimizer
and diagnostic programs. The policy conditions these decisions; the model, data, loss, and update budget remain
fixed. Final metrics are withheld from the in-run agent. Held-out tasks are reserved for transfer evaluation;
\Cref{app:audit} documents the access protocol and a historical exposure during evolution.}
\label{fig:overview}
\end{figure}

Neural-network training usually begins with an optimizer and schedule chosen in advance. Their effectiveness can
depend on the architecture, data, and training dynamics, yet selecting a new recipe often requires repeated runs
and manual intervention. A coding agent offers another form of adaptation: it can inspect a training run and revise
the optimizer's program while the model trains.\\
Concurrent work on reactive optimizers, \opal{}~\citep{bai2026opalreactive}, studies this within-run setting. A
single LLM agent observes training dynamics and revises the update rule, learning rates, schedules, and diagnostics
during an individual run. Its conversation persists within that run, but its fixed initial policy does not retain
lessons for the next one. A useful optimizer change, a misleading diagnostic, or a tested implementation must be
rediscovered when a new run starts.We study whether optimizer agents can improve this initial policy across runs through recursive self-improvement.
\method{} makes the agent's accumulated knowledge an explicit, editable object: a prompt, a playbook of procedures
and lessons, a library of optimizer and diagnostic code, and a decision cadence. An outer meta-agent reads evidence
from earlier runs, modifies a parent policy, and evaluates the child on a training suite. An archive supports
continued exploration and inheritance, following the archive-based approach of the Darwin G\"odel
Machine~\citep{zhang2025dgm}. The underlying LLM weights, meta-agent instructions, and evaluator remain fixed;
the policy used to improve optimizers is the object that evolves (\Cref{fig:overview}). This separation makes the learned knowledge inspectable, but it also exposes a generalization question. A policy
can improve its evolution score by accumulating broadly useful procedures or by fitting recipes to a few familiar
tasks. Better training-suite performance alone does not distinguish the two. Nor does a growing code library show
whether an agent still needs to adapt during a new run. We therefore examine the content of the evolved policies,
their transfer to unseen tasks and other LLMs, and the contribution of decisions made after the initial optimizer
change. Our contributions are:
\begin{itemize}
\item \textbf{An explicit policy-evolution framework for optimizer agents.} \method{} adds cross-run policy
evolution to the within-run optimizer-agent setting. Our experiments connect a coding agent to an editable optimizer interface, allowing it to monitor training dynamics and revise update rules during a run.
fixed baseline prompt, and four example tasks, with attribution and implementation details in
\Cref{sec:setup}. The contribution is the evolving policy, outer search, and study of the resulting knowledge.
\item \textbf{An evaluation of transfer and online adaptation.} We evaluate nine single-GPU tasks with three
in-run LLMs, fresh evaluation seeds, comparisons of reported means, and controls using an unedited library template, a frozen
first optimizer change, a one-decision agent, and the full agent. Three tasks guide evolution; three ML tasks and
three PDE problems test transfer beyond that suite.
\item \textbf{Evidence of selective transfer and specialization.} Evolved policies improve the training suite
for all three LLMs. GPT-5.5 and GPT-6.1 improve the mean summary score in every task group;
Opus~5.5 improves on four of six held-out tasks but has a lower aggregate held-out summary score. Initial adaptation captures most of the observed benefit on
seen tasks, while later decisions improve unseen-task performance for both baseline and evolved agents. Held-out
ML transfer peaks early as continued evolution increasingly fits the training suite.
\end{itemize}

The meta-agents accumulate substantive artifacts: reusable update modules, parameter-role checks, scheduling
procedures, and CPU proxy experiments. We analyze these artifacts alongside their measured outcomes rather than
treating higher archive scores as sufficient evidence of general improvement. The harness, policies, transcripts,
and results are released.

\section{Related work}
\label{sec:related}

\paragraph{Learned and discovered optimizers.} Learned optimizers~\citep{andrychowicz2016learning,metz2022velo}
meta-train an update rule across tasks, while symbolic search has discovered update algorithms such as
Lion~\citep{chen2023lion}. The resulting update program is typically fixed before a target training run, although
its updates can depend on the current state. In \method{}, the evolved object is the optimizer agent's policy,
which guides further program edits during each target run. Established optimizers such as
Adam~\citep{kingma2015adam}, AdamW~\citep{loshchilov2019adamw}, Shampoo~\citep{gupta2018shampoo},
SOAP~\citep{vyas2024soap}, and Muon~\citep{jordan2024muon}, together with recent extensions such as spectral orthogonal gradient projection for continual learning~\citep{lu2026muon}, supply building blocks that agents can select and compose.

\paragraph{Agents that improve programs and policies.} LLMs have been used to optimize prompts and
solutions~\citep{yang2023opro}, evolve prompts~\citep{fernando2023promptbreeder}, improve code-generation
scaffolds~\citep{zelikman2023stop}, design agentic systems~\citep{hu2024adas}, and evolve algorithms against an
evaluator~\citep{novikov2025alphaevolve}. The Darwin G\"odel Machine~\citep{zhang2025dgm} maintains an archive of
modified coding agents and samples parents for further exploration, drawing on the idea of a self-improving
G\"odel machine~\citep{schmidhuber2003godel}. Our outer search adopts this archive-based structure. Persistent
natural-language lessons~\citep{shinn2023reflexion}, skill libraries~\citep{wang2023voyager}, and self-evolving procedural memory learned from contrasting multi-model trajectories~\citep{deng2026contramem} provide related
mechanisms for carrying experience across episodes. \method{} combines procedures and executable tools in an
optimizer-agent policy. Every valid evaluated child enters the archive; scores guide subsequent parent and final
policy selection rather than acting as a strict improvement filter.

\paragraph{Optimizer agents within a training run.} We found concurrent work, OPAL, introduces a coding agent that monitors training dynamics and revises optimizer and diagnostic programs within a single run. Each new run starts from a fixed prompt, without retaining experience from previous runs. RSI-Opt addresses the complementary problem of improving the agent’s policy across runs by accumulating and refining optimization procedures and executable tools. We examine how this knowledge transfers to unseen tasks and other LLMs, and when continued evolution leads to specialization. Population-based training~\citep{jaderberg2017pbt}  also adapts training hyperparameters through selection and replacement across a population of runs.

\paragraph{Optimization for physics-informed training.} PINN training can suffer from imbalanced gradients across
residual and boundary losses~\citep{wang2021gradient} and ill-conditioned loss landscapes, motivating approaches
such as Adam followed by L-BFGS and other second-order methods~\citep{rathore2024challenges}, as well as adaptive optimizers for hard-constrained PINN training~\citep{lu2026adamflip}. Our PDE problems test
transfer beyond the ML tasks used for evolution. They assess the applicability of evolved optimizer-agent
knowledge; they do not establish a new PDE discretization or a universally better PINN optimizer.

\section{Method}
\label{sec:method}

\subsection{Within-run adaptation}
The inner loop exposes the optimizer’s update rule as an editable program, allowing a coding agent to inspect training losses and diagnostic reports and revise the optimization procedure at scheduled decision points. The training script, model architecture, data order, loss, batch size, and
number of parameter updates are fixed. At $D$ scheduled decisions, the agent receives the training-loss history
since its previous decision, reports from its analysis program, errors, and timing information. It can inspect and
edit files in its workspace and run shell commands without network or GPU access. The agent edits two programs. coder receives gradients at each update and modifies the parameters in
place. Analyst computes read-only diagnostics every updates. A new updater is tested
over a short trial window and rolled back if it raises an exception. Training pauses while the agent decides, so
agent latency affects wall-clock cost rather than the number of model updates. Conversation history persists
across decisions within the run. The final evaluation metric is withheld from the in-run agent and used by the
outer evaluator after the run. Our \emph{\baseline{}} uses  fixed prompt without an accumulated playbook or library. This prompt supplies
the task description, protocol, and program interfaces. The same template is instantiated for each run, and
experience from completed runs is not retained in the next run's policy.

\subsection{The evolving policy}
A policy $\pi$ specifies what the in-run agent brings to a run:
\begin{itemize}
\item \textbf{Prompt template:} instructions with task-specific fields for the goal, parameter table, model
summary, starting optimizer, and protocol.
\item \textbf{Playbook}: procedures and lessons written in natural language.
\item \textbf{Code library}: indexed Python modules for update rules, schedules, parameter
grouping, stability checks, and diagnostics.
\item \textbf{Cadence:} a front-loaded or uniform decision schedule and the frequency of diagnostic checks.
\end{itemize}
The playbook and library are read-only references during a run; the agent can copy a library module into its
workspace and use or modify the copy. The experiment fixes the number of decision opportunities rather than
allowing the policy to increase it. Equal decision limits and cost caps do not imply equal realized token use,
diagnostic overhead, or training time. Generation~0 uses the baseline prompt with one additional paragraph pointing
to an empty playbook and library.

\subsection{Across-run policy evolution}
The outer loop maintains an archive of evaluated policies, following the archive-based search of the Darwin
G\"odel Machine~\citep{zhang2025dgm}. To generate a child, it samples an eligible parent $\pi_p$ with probability
proportional to
\[
\frac{\operatorname{sigmoid}\!\left(\lambda(s_p-\widetilde{s})\right)}{1+n_p},
\]
where $s_p$ is the parent's score, $\widetilde{s}$ is the archive median score, and $n_p$ counts its children. The
score term favors stronger parents, while the child count reduces repeated selection of the same branch.\\
A separate \emph{meta-agent} receives a copy of the parent policy and digests of its evaluation runs. Each digest
contains the final metric and z-score, the in-run agent's recorded rationales, final optimizer program, and notes.
The meta-agent also sees archive entries with per-task scores and changelogs. It edits the policy and is instructed
to record its changes, supporting evidence, and expected effects. Its CPU-only shell permits unit tests and proxy
experiments. The instructions encourage transferable lessons and prohibit task names and task-specific branches;
this instruction does not itself prevent specialization. Validation checks required template fields, Python parsing, and size limits. A child that fails these checks is
returned to the meta-agent for repair, at most twice. Every valid child is evaluated on the training suite and added
to the archive, including children that score below their parent. Thus the archive retains explored alternatives,
while performance guides further parent selection and the final policy choice.

\paragraph{Evaluation score.} For task $t$, let $h_t(m)=m$ for ML metrics and $h_t(m)=\log_{10}m$ for PDE solution
errors. We estimate the static recipe's mean $\mu_t$ and standard deviation $\sigma_t$ from five seeds in this
transformed metric space. A policy run with final metric $m_{t,s}$ receives
\[
z_{t,s}=d_t\,\frac{h_t(m_{t,s})-\mu_t}{\sigma_t},
\qquad
d_t=\begin{cases}+1,&\text{higher is better},\\-1,&\text{lower is better}.\end{cases}
\]
Positive values indicate improvement over the static reference. The policy score is
\[
s(\pi)=\frac{1}{|\mathcal{T}|}\sum_{t\in\mathcal{T}}
       \frac{1}{|S|}\sum_{s\in S}z_{t,s}.
\]
Policies use the same task and seed set during evolution. This normalization is an aggregation rule, not a test of
statistical significance; tasks with small static variability can contribute large z-scores.

\paragraph{Final descriptive summaries.} To make final aggregate comparisons reproducible from
\Cref{tab:main}, we separately report raw-space summary scores
\[
\bar z_t^{\mathrm{raw}}=d_t\,\frac{\bar m_t-\bar m_t^{\mathrm{static}}}{s_t^{\mathrm{static}}},
\qquad
\bar z_{\mathcal G}^{\mathrm{raw}}=\frac{1}{|\mathcal G|}\sum_{t\in\mathcal G}\bar z_t^{\mathrm{raw}},
\]
using the printed mean metrics and static standard deviations for every task, including PDEs. These descriptive
scores do not change policy selection, which uses only the ML training suite. They differ from PDE scores computed
from per-seed log errors and have limited precision because the table entries are rounded. Final aggregate scores
and differences in \Cref{tab:agg,tab:paired} and the accompanying prose use this raw-space summary convention.
Individual-seed confidence intervals and seed-win counts cannot be reconstructed from means and standard
deviations alone and are not reported in the corrected comparison.

\paragraph{Scope of recursive self-improvement.} The meta-agent improves the policy by which the in-run agent
improves optimizers. Later generations inherit and revise the procedures and tools accumulated by earlier ones.
The underlying LLM weights, meta-agent instructions, scoring rule, and designated task splits remain fixed. This
is recursive improvement of an agent policy, without modification of the meta-agent's own improvement procedure.
The experiments measure the resulting policies; a separate matched-budget control would be needed to isolate the
benefit of inheritance over independent proposals.

\begin{algorithm}[t]
\caption{\method: policy evolution}
\label{alg:rsi}
\begin{algorithmic}[1]
\Require Training suite $\mathcal{T}$, seeds $S$, static reference statistics, generations $G$, children $K$
\State Initialize $\pi_0$ from the baseline prompt with empty playbook and library
\State Evaluate $\pi_0$ on $\mathcal{T}\times S$; initialize archive $\mathcal{A}\gets\{(\pi_0,s_0)\}$
\For{$g=1,\ldots,G$}
  \For{$k=1,\ldots,K$ \textbf{in parallel}}
    \State Sample an eligible parent from $\mathcal{A}$ using the score/child-count distribution
    \State Propose $\pi_c\gets\textsc{MetaAgent}(\pi_p,\text{run digests},\mathcal{A})$
    \State Validate $\pi_c$ and allow at most two repairs
    \If{$\pi_c$ is valid}
      \State Run $\pi_c$ on $\mathcal{T}\times S$ and compute $s_c$
      \State Add $(\pi_c,s_c)$ and its run evidence to $\mathcal{A}$
    \EndIf
  \EndFor
\EndFor
\State Select a final policy using training-suite scores and the re-evaluation rule in \Cref{sec:setup}
\State Evaluate the selected policy on fresh seeds and held-out tasks
\end{algorithmic}
\end{algorithm}

\subsection{Implementation}
Both the in-run agent and the meta-agent operate through a tool-calling loop on the API, with access to four tools: \texttt{bash}, \texttt{read\_file}, \texttt{write\_file}, and \texttt{edit\_file}. The backend is self-contained and leaves the inner-loop training code untouched. For isolation, each agent's shell executes in its own network namespace with no access to GPUs or API credentials. When a conversation grows beyond 160k tokens, the agent writes a summary of its progress, which then replaces the earlier history. Runs are scheduled across GPUs through a file-based queue, and because training halts whenever the agent is deciding, multiple runs can share a single GPU efficiently. Every LLM is configured with high reasoning effort. Filesystem restrictions and the extent of their coverage are detailed in \Cref{app:audit}.
\section{Experimental setup}
\label{sec:setup}

We compare baseline \baseline{}  that studies a coding agent that monitors a single training run and revises its optimizer as that run unfolds At generation~0, the only change is one added paragraph pointing the agent to the policy folder, which starts empty. The contributions of \method{} are the policy representation and the procedure that evolves it across runs.

\paragraph{Tasks.} We evaluate nine single-GPU training tasks in three groups (\Cref{tab:tasks}). Each task has a
fixed model, loss, data protocol, update budget, and starting optimizer. The agent can change optimization but not
these task definitions.
\begin{itemize}
\item \textbf{Training suite:} ResNet-18~\citep{he2016resnet} on
CIFAR-10~\citep{krizhevsky2009cifar} with SGD-Nesterov and a one-cycle schedule; a 34M-parameter GPT on
FineWeb~\citep{penedo2024fineweb} with AdamW and warmup-cosine decay; and a small UNet trained by flow
matching~\citep{lipman2023flow} on MNIST with AdamW, cosine decay, and EMA. These tasks provide the feedback used
by the meta-agent during evolution.
\item \textbf{Held-out ML:} a ViT~\citep{dosovitskiy2021vit} on CIFAR-100 with AdamW and warmup-cosine decay; a
2-layer LSTM on WikiText-2~\citep{merity2017wikitext} with Adam and step decay; and \opal{}'s noisy-label CIFAR-10
task with 10\% incorrect labels. In the last task, the agent sees a development set but not test accuracy.
\item \textbf{Held-out PDE:} three existing neural solvers with their models, sampling, losses, and update budgets
unchanged. They are a PINN~\citep{raissi2019pinn} for viscous Burgers in 1D with full-batch Adam and 10k updates;
a PINN for heat diffusion in 2D with resampled collocation points and 30k Adam updates; and a variable-order
time-fractional diffusion solver using finite-difference discretization and full-batch Adam for 10k updates,
related to fractional physics-informed training~\citep{pang2019fpinns}. Each is evaluated by relative $L_2$
solution error.
\end{itemize}
Held-out tasks are designated for transfer evaluation rather than policy selection. \Cref{app:audit} documents an
accidental read of one held-out task description during evolution and the subsequent exclusions. The PDE tasks
differ from the evolution suite in architecture, residual-based objectives, sampling regime, and evaluation metric.

\begin{table}[t]
\centering
\small
\caption{Tasks. Static: the task's own recipe, mean $\pm$ std over 5 seeds of the final metric.}
\label{tab:tasks}
\begin{tabular}{llllr}
\toprule
Group & Task & Static recipe & Metric & Static result \\
\midrule
Training suite & ResNet-18 / CIFAR-10 & SGD-Nesterov, one-cycle & test acc. $\uparrow$ & 94.60{\scriptsize$\pm$0.16} \\
 & GPT-34M / FineWeb & AdamW, warmup-cosine & val. loss $\downarrow$ & 4.480{\scriptsize$\pm$0.008} \\
 & Flow UNet / MNIST & AdamW, cosine, EMA & EMA val. MSE $\downarrow$ & 0.1700{\scriptsize$\pm$0.0004} \\
\midrule
Held-out ML & ViT / CIFAR-100 & AdamW, warmup-cosine & test acc. $\uparrow$ & 55.84{\scriptsize$\pm$0.34} \\
 & LSTM / WikiText-2 & Adam, step decay & test loss $\downarrow$ & 5.386{\scriptsize$\pm$0.064} \\
 & ResNet-18 / noisy CIFAR-10 & AdamW, warmup-cosine & test acc. $\uparrow$ & 73.76{\scriptsize$\pm$1.71} \\
\midrule
Held-out PDE & PINN / Burgers 1D & Adam (full batch) & rel. $L_2$ $\downarrow$ & 0.0409{\scriptsize$\pm$0.013} \\
 & PINN / Heat 2D & Adam (resampled points) & rel. $L_2$ $\downarrow$ & 0.378{\scriptsize$\pm$0.36} \\
 & Fractional PDE & Adam (full batch) & rel. $L_2$ $\downarrow$ & 0.384{\scriptsize$\pm$0.061} \\
\bottomrule
\end{tabular}
\end{table}

\paragraph{Models.} We use Claude Opus~5.5, GPT-5.5, and GPT-6.1 as in-run agents, all with
high reasoning effort. Both the meta-agent and in-run agent use Opus~5.5 during evolution. The resulting policy is
then used unchanged with each in-run model.

\paragraph{Baselines and budget controls.} The \textbf{static} reference is each script's original optimizer and
schedule, evaluated on five seeds. The \textbf{tuned-static} reference multiplies its learning rate by
$\{\frac14,\frac12,1,2,4\}$, with three seeds per multiplier. The best multiplier is selected on the evaluation
metric itself; this is an optimistic oracle comparison, not a deployable tuning protocol. The
\textbf{\baseline{}} is  fixed-prompt agent without an accumulated playbook or library. Full baseline
and \method{} runs share a backend, \decisionsPerRun{} decision opportunities, the available schedule families,
and a per-run cost cap. Cadence and realized resource use can differ. \Cref{sec:res-budget} additionally compares
an unedited evolved-library recipe, a frozen first optimizer change, and a one-decision agent.

\paragraph{Policy selection and final evaluation.} Evolution evaluates each policy on the training suite using
seeds 100 and 101. Before final evaluation, the three best eligible policies by archive score are re-evaluated on
seeds 102 and 103, and the highest four-seed mean determines the final policy. Reported final comparisons use fresh
seeds 200, 201, and 202, which do not guide evolution or final-policy selection. Training-suite metrics guide the
search even when a task script names its evaluation split ``test''; fresh seeds on these tasks assess performance
within the seen suite, not generalization to unseen tasks. We report raw metrics as mean $\pm$ standard deviation across seeds. Final aggregate comparisons use the raw-space
summary scores defined in \Cref{sec:method}, recomputed from the printed statistics in \Cref{tab:main}.
The corrected difference table reports signed differences of these means; it does not infer paired confidence
intervals or individual-seed win counts from summary statistics. We also evaluate the best eligible policies after generations 1 and 3 on the same fresh seeds. This trajectory
analysis measures transfer over evolution; held-out outcomes are not used to replace the preselected final policy.

\section{Results}
\label{sec:results}

\subsection{What knowledge evolves?}
\label{sec:res-evolution}

\begin{figure}[t]
\centering
\includegraphics[width=\linewidth]{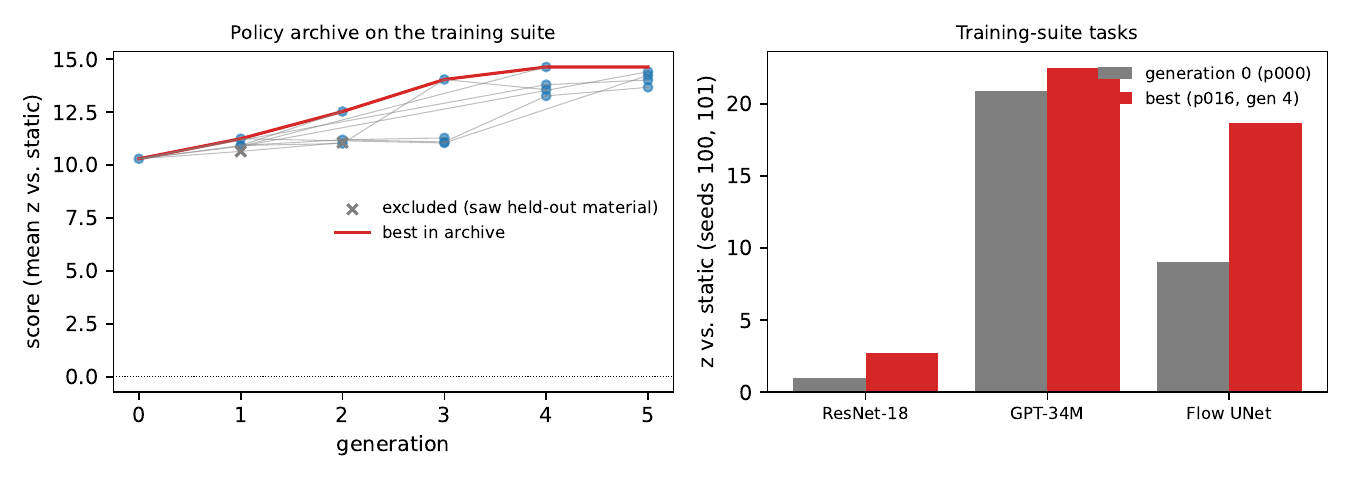}
\caption{Left: policies evaluated during evolution on training-suite seeds 100 and 101. Gray lines connect
children to parents; the red line shows the best archive score so far. Right: per-task z-scores of generation~0 and
the selected final policy on these evolution seeds. Fresh-seed comparisons are reported separately.}
\label{fig:evolution}
\end{figure}

Evolution produced \nPolicies{} policies over \nGenerations{} generations of four children, with no invalid policies
(\Cref{fig:evolution}). Generation~0, which augments the fixed prompt with references to an empty playbook and
library, scored \genZeroScore{}. The best archive score reached \bestScore{} for policy \bestPolicy{} in
generation~\bestGen{}. Re-evaluating the three highest-scoring eligible candidates on two additional seeds selected
\bestPolicy{}: its four-seed mean was \scoreFourPSixteen{}, compared with \scoreFourPNineteen{} and
\scoreFourPTwenty{} for the other candidates. These archive scores guide search and selection; the remaining
comparisons use fresh seeds.

\paragraph{Procedures and executable tools.} The meta-agents changed the fixed prompt only modestly: children
added 0--23 lines, mainly summaries pointing to the playbook. Most accumulated knowledge appeared in procedures and
library modules. Early policies recorded useful changes, including Muon on transformer matrices, learning-rate
decay to zero, and parameter-role-specific treatment. They also warned against interpreting a transient loss drop
after a rate reduction as durable progress or triggering changes from noisy window means. The resulting library included a hybrid updater using Muon on selected tensors and the starting rule elsewhere,
schedule helpers, a parameter-role classifier, diagnostics, and interface checks. A recurring procedure was to make
an early, evidence-based change and retain it unless later observations justified another edit. Nine of the 20
meta-agents, including the one that produced \bestPolicy{}, reached the 50-call or 30-minute limit before writing
a changelog entry. Their edits were retained, so changelog coverage is incomplete even though every child differs
from its parent.

\paragraph{A proxy-supported optimizer change.} The largest observed task-score jump followed a hypothesis that
a meta-agent tested using its own CPU proxy. It trained a small GroupNorm UNet by flow matching from a
constant-rate AdamW baseline. Adding Muon on hidden convolutions reduced validation loss in every configuration it
tested, including 0.188 to 0.174 at 3000 steps. Within that proxy sweep, the learning rate on the remaining
Adam-updated tensors mattered more than Muon's own rate. The meta-agent recorded the proposal as an untested
transfer from the proxy to the actual task. Subsequent in-run agents applied it, and the UNet's evolution score
rose from about $+9$ to $+19$ on both seeds in generation~3. The proposal also appeared in other lineages after
their meta-agents read archive summaries. This sequence connects a recorded hypothesis, executable tests, and a
later gain; isolating the change's causal contribution would require a controlled parent-to-child replay.

\paragraph{Exposure and exclusions.} Before filesystem restrictions were added, one generation-1 meta-agent read
 description of its noisy-label example, a designated held-out task. That policy and its descendant
(\nExcluded{} policies) were excluded from parent selection and final selection. The audit identified no additional
direct reads of held-out material. Because archive summaries can transmit information across branches, direct-read
and ancestry checks alone do not establish the absence of indirect exposure (\Cref{app:audit}).

\subsection{How do agents compare with static recipes?}
\label{sec:res-baselines}

\providecommand{\pmstd}[1]{{\scriptsize$\pm$#1}}

\begin{table*}[t]
\centering
\small                          
\setlength{\tabcolsep}{5pt} 
\caption{Final metric on fresh seeds (mean $\pm$ std over 3 seeds; static: 5 seeds). For each in-run model, the better of the \baseline{} and \method{} is in bold. Accuracies in \%. The training-suite rows are the tasks the policy was evolved on; all other rows are held out.}
\label{tab:main}
\begin{adjustbox}{max width=\textwidth}
\begin{tabular}{lllcccccc}
\toprule
 & & & \multicolumn{2}{c}{Opus 5.5} & \multicolumn{2}{c}{GPT-5.5} & \multicolumn{2}{c}{GPT-6.1} \\
\cmidrule(lr){4-5} \cmidrule(lr){6-7} \cmidrule(lr){8-9}
Task & Static & Tuned LR & \Baseline{} & \method{} & \Baseline{} & \method{} & \Baseline{} & \method{} \\
\midrule
\multicolumn{9}{l}{\textit{Training suite}} \\
ResNet-18 / CIFAR-10 & 94.60\pmstd{0.16} & 94.62\pmstd{0.23}{\tiny\,$\times$1} & 94.55\pmstd{0.23} & \textbf{94.78\pmstd{0.03}} & 94.58\pmstd{0.17} & \textbf{94.77\pmstd{0.17}} & 94.41\pmstd{0.06} & \textbf{94.86\pmstd{0.11}} \\
GPT-34M / FineWeb & 4.480\pmstd{0.008} & 4.478\pmstd{0.003}{\tiny\,$\times$1} & 4.332\pmstd{0.018} & \textbf{4.307\pmstd{0.007}} & 4.407\pmstd{0.098} & \textbf{4.308\pmstd{0.008}} & 4.315\pmstd{0.011} & \textbf{4.310\pmstd{0.005}} \\
Flow UNet / MNIST & 0.1700\pmstd{0.0004} & 0.1660\pmstd{0.0003}{\tiny\,$\times$4} & 0.1647\pmstd{0.0021} & \textbf{0.1619\pmstd{0.0003}} & 0.1684\pmstd{0.0007} & \textbf{0.1621\pmstd{0.0001}} & 0.1659\pmstd{0.0001} & \textbf{0.1621\pmstd{0.0004}} \\
\midrule
\multicolumn{9}{l}{\textit{Held-out ML}} \\
ViT / CIFAR-100 & 55.84\pmstd{0.34} & 55.81\pmstd{0.47}{\tiny\,$\times$1} & \textbf{62.62\pmstd{1.15}} & 61.52\pmstd{1.30} & 57.56\pmstd{3.63} & \textbf{61.98\pmstd{0.19}} & 55.54\pmstd{1.39} & \textbf{61.92\pmstd{0.30}} \\
LSTM / WikiText-2 & 5.386\pmstd{0.064} & 5.296\pmstd{0.016}{\tiny\,$\times$0.5} & 5.176\pmstd{0.049} & \textbf{5.175\pmstd{0.003}} & \textbf{5.437\pmstd{0.092}} & 5.467\pmstd{0.065} & \textbf{5.315\pmstd{0.052}} & 5.439\pmstd{0.062} \\
ResNet-18 / noisy CIFAR-10 & 73.76\pmstd{1.71} & 74.02\pmstd{0.97}{\tiny\,$\times$0.5} & 78.73\pmstd{3.50} & \textbf{81.77\pmstd{1.99}} & 72.52\pmstd{7.12} & \textbf{78.58\pmstd{0.85}} & 73.30\pmstd{1.98} & \textbf{79.82\pmstd{0.39}} \\
\midrule
\multicolumn{9}{l}{\textit{Held-out PDE}} \\
PINN / Burgers 1D & 4.09e-2\pmstd{1.3e-2} & 2.28e-2\pmstd{2.1e-2}{\tiny\,$\times$4} & 3.95e-3\pmstd{3.4e-3} & \textbf{1.45e-3\pmstd{1.5e-3}} & 1.41e-1\pmstd{5e-2} & \textbf{8.54e-2\pmstd{7.3e-2}} & 4.82e-3\pmstd{4.5e-3} &  \textbf{5.46e-4\pmstd{4.4e-4}}\\
PINN / Heat 2D & 3.78e-1\pmstd{3.6e-1} & 7.93e-3\pmstd{6.2e-3}{\tiny\,$\times$0.25} &6.3e-5\pmstd{3.6e-5}  &  \textbf{2.9e-5\pmstd{1.2e-5}}& 3.55e-3\pmstd{3.6e-3} & \textbf{2.37e-4\pmstd{2.8e-4}} &3.27e-5\pmstd{4e-6} &  \textbf{1.6e-5\pmstd{3.9e-6}} \\
Fractional PDE & 0.384\pmstd{0.061} & 0.329\pmstd{0.028}{\tiny\,$\times$0.5} & \textbf{0.306\pmstd{0.095}} & 0.357\pmstd{0.069} & 0.326\pmstd{0.022} & \textbf{0.297\pmstd{0.015}} &0.322\pmstd{0.068} &  \textbf{0.319\pmstd{0.002}} \\
\bottomrule
\end{tabular}
\end{adjustbox}
\end{table*}

\Cref{tab:main} reports final raw metrics. With Opus~5.5, the \baseline{} improves GPT validation loss from 4.480
to 4.332 and ViT accuracy from 55.8\% to 62.6\%. Its mean heat-equation error is approximately 6,000 times
lower than the mean error of the static recipe, whose constant learning rate leaves a noisy final iterate. This large
comparison is an agent-versus-static result, not an incremental improvement of \method{} over the baseline.

The oracle learning-rate sweep recovers part of these gains: its mean held-out score is \zTunedHeld{}, compared
with \zOpalOpusHeld{} for the Opus baseline. Benefits are not universal. For example, the GPT-5.5 baseline worsens
both Burgers error and LSTM loss relative to their static recipes. The results therefore depend on both the task
and the LLM controlling the optimizer.

\subsection{What does the evolved policy transfer?}
\label{sec:res-transfer}

\begin{table}[t]
\centering\small
\setlength{\tabcolsep}{4pt}
\caption{Task-balanced raw-space summary $z$-scores recomputed from the printed means and static standard
deviations in \Cref{tab:main}; positive is better than the static recipe. Wins count tasks on which \method{}
has the better reported mean, not individual seeds. Scores are descriptive and subject to input rounding.}
\label{tab:agg}
\begin{tabular}{llrrrr}
\toprule
Model & Method & Train & Held-out ML & Held-out PDE & All held-out \\
\midrule
-- & Static, tuned LR (oracle) & \zTunedTrain{} & \zTunedHeldML{} & \zTunedHeldPDE{} & \zTunedHeld{} \\
Opus 5.5 & \Baseline{} & \zOpalOpusTrain{} & \zOpalOpusHeldML{} & \zOpalOpusHeldPDE{} & \zOpalOpusHeld{} \\
 & \method{} & \zRsiOpusTrain{} & \zRsiOpusHeldML{} & \zRsiOpusHeldPDE{} & \zRsiOpusHeld{} \\
 & wins & 3/3 & 2/3 & 2/3 & 4/6 \\
GPT-5.5 & \Baseline{} & \zOpalGptFiveTrain{} & \zOpalGptFiveHeldML{} & \zOpalGptFiveHeldPDE{} & \zOpalGptFiveHeld{} \\
 & \method{} & \zRsiGptFiveTrain{} & \zRsiGptFiveHeldML{} & \zRsiGptFiveHeldPDE{} & \zRsiGptFiveHeld{} \\
 & wins & 3/3 & 2/3 & 3/3 & 5/6 \\
GPT-6.1 & \Baseline{} & \zOpalGptSixTrain{} & \zOpalGptSixHeldML{} & \zOpalGptSixHeldPDE{} & \zOpalGptSixHeld{} \\
 & \method{} & \zRsiGptSixTrain{} & \zRsiGptSixHeldML{} & \zRsiGptSixHeldPDE{} & \zRsiGptSixHeld{} \\
 & wins & 3/3 & 2/3 & 3/3 & 5/6 \\
\bottomrule
\end{tabular}
\end{table}

\begin{table}[t]
\centering\small
\setlength{\tabcolsep}{5pt}
\caption{Signed differences between the reported mean metrics in \Cref{tab:main}, positive when \method{}
is better. Task rows use accuracy points, loss units, or raw relative $L_2$ error; PDE rows are not differences
of per-seed log errors. Group rows are differences of the raw-space summary $z$-scores in \Cref{tab:agg},
computed before rounding. Individual-seed win counts and paired bootstrap intervals are not inferred from
aggregate means and standard deviations.}
\label{tab:paired}
\begin{tabular}{lrrr}
\toprule
Task & Opus 5.5 & GPT-5.5 & GPT-6.1 \\
\midrule
ResNet-18 / CIFAR-10 & +0.23 & +0.19 & +0.45 \\
GPT-34M / FineWeb & +0.025 & +0.099 & +0.005 \\
Flow UNet / MNIST & +0.0028 & +0.0063 & +0.0038 \\
\textit{Training suite: mean $\Delta\bar z^{\mathrm{raw}}$} & \dzOpusTrain{} & \dzGptFiveTrain{} & \dzGptSixTrain{} \\
\midrule
ViT / CIFAR-100 & -1.10 & +4.42 & +6.38 \\
LSTM / WikiText-2 & +0.001 & -0.030 & -0.124 \\
ResNet-18 / noisy CIFAR-10 & +3.04 & +6.06 & +6.52 \\
\textit{Held-out ML: mean $\Delta\bar z^{\mathrm{raw}}$} & \dzOpusHeldML{} & \dzGptFiveHeldML{} & \dzGptSixHeldML{} \\
\midrule
PINN / Burgers 1D & $+2.50\!\times\!10^{-3}$ & $+5.56\!\times\!10^{-2}$ & $+4.27\!\times\!10^{-3}$ \\
PINN / Heat 2D & $+3.40\!\times\!10^{-5}$ & $+3.31\!\times\!10^{-3}$ & $+1.67\!\times\!10^{-5}$ \\
Fractional PDE & -0.051 & +0.029 & +0.003 \\
\textit{Held-out PDE: mean $\Delta\bar z^{\mathrm{raw}}$} & \dzOpusHeldPDE{} & \dzGptFiveHeldPDE{} & \dzGptSixHeldPDE{} \\
\bottomrule
\end{tabular}
\end{table}

\paragraph{New seeds on seen tasks.} With Opus~5.5, \method{} raises the training-suite score from
\zOpalOpusTrain{} to \zRsiOpusTrain{}, a summary-score difference of \dzOpusTrain{}.
Fresh seeds show that the selected policy's training-suite gain is not restricted to the seeds
used for search. The tasks themselves remain those used during evolution.

\paragraph{Unseen tasks with the evolution model.} On the six held-out tasks, Opus has a lower aggregate summary score:
\zOpalOpusHeld{} for the baseline versus \zRsiOpusHeld{} for \method{}. The held-out ML and PDE summary-score
differences are \dzOpusHeldML{} and \dzOpusHeldPDE{}, respectively. Per-task effects are mixed:
\method{} reduces mean Burgers error by about $2.7\times$, reduces heat error by about $2.2\times$, and adds
approximately 3 accuracy points on noisy CIFAR-10, but loses 1.1 points on the ViT and increases fractional-PDE error.
It improves the mean metric on four of the six held-out tasks; this count need not agree with the direction of
the standardized aggregate.

\paragraph{Other in-run models.} The same policy, used unchanged with GPT-5.5, raises its mean held-out score from
\zOpalGptFiveHeld{} to \zRsiGptFiveHeld{}. It also brings the training-suite score close to Opus with the same
policy (\zRsiGptFiveTrain{} versus \zRsiOpusTrain{}). Its held-out ML and PDE summary-score differences are
\dzGptFiveHeldML{} and \dzGptFiveHeldPDE{}, respectively.

GPT-6.1 improves the aggregate summary score in every task group. Its training-suite difference is
\dzGptSixTrain{}, and its held-out ML difference is \dzGptSixHeldML{}, including approximately 6 additional
accuracy points on each image-classification task. Its PDE difference is positive: \dzGptSixHeldPDE{}.
On Burgers, mean relative error falls from $4.82\times10^{-3}$ to $5.46\times10^{-4}$, about $8.8$ times lower.
Mean heat error falls from $3.27\times10^{-5}$ to $1.60\times10^{-5}$, about $2.0$ times lower, while
fractional-PDE error changes little ($0.322$ to $0.319$). Its overall held-out summary score rises from
\zOpalGptSixHeld{} to \zRsiGptSixHeld{}, although its LSTM loss worsens.

These results suggest that portable knowledge helps when it compensates for a model's weaker task-specific
choices, but can interfere with an already effective strategy, consistent with evidence that different LLMs hold complementary knowledge~\citep{cao2026medrouter}. They do not support a uniform improvement across
models or domains. Library use by the GPT agents (\Cref{tab:behavior}) shows that the artifact is adopted; its
performance determines whether that adoption is beneficial.

\subsection{When does evolution specialize?}
\label{sec:res-specialize}

\begin{figure}[t]
\centering
\includegraphics[width=0.75\linewidth]{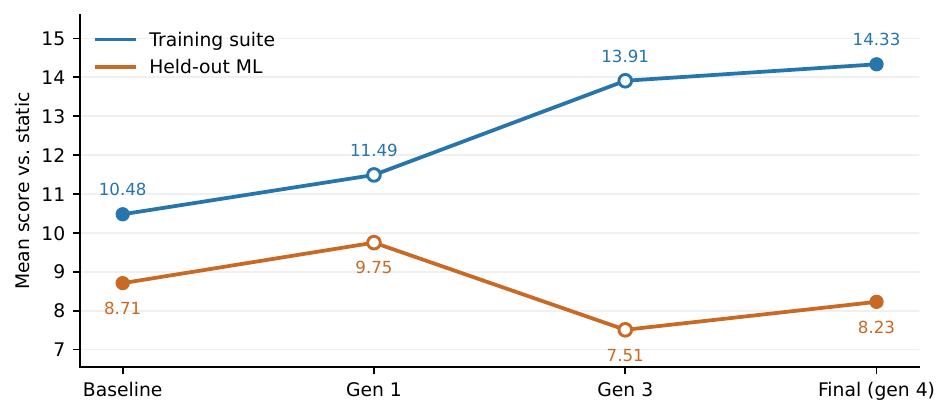}
\caption{Opus training-suite and held-out ML score trajectories. Baseline and final endpoints are raw-space
summary scores recomputed from \Cref{tab:main}; intermediate checkpoints retain their originally reported scores
and precision. Policies are selected using training-suite scores, not the held-out outcomes shown here.
PDE checkpoint aggregates are omitted because their per-task raw metrics are unavailable for this recomputation.}
\label{fig:trajectory}
\end{figure}

The training-suite score increases at each evaluated checkpoint
(\trajOpalTrain{} $\to$ \trajGenOneTrain{} $\to$ \trajGenThreeTrain{} $\to$ \zRsiOpusTrain{}), but held-out ML
performance peaks after the first generation
(\trajOpalHeldML{} $\to$ \trajGenOneHeldML{} $\to$ \trajGenThreeHeldML{} $\to$ \zRsiOpusHeldML{};
\Cref{fig:trajectory}). Baseline and final values are recomputed from \Cref{tab:main}; intermediate values retain
their original reporting precision. Thus an early transfer peak is supported for held-out ML; the corrected
trajectory comparison does not establish the ordering of PDE checkpoints.

Generation~1 mainly added general procedures: choose an early change carefully, decay the learning rate, and check
tensor roles. Later policies increasingly documented measured recipes for the three evolution tasks. For example,
the playbook's table grew from 6 to 18 runs, and the convolutional Muon proposal targeted the UNet. These artifacts
are consistent with specialization to the small training suite. A broader evolution suite or a separate policy
validation suite could help, at additional evaluation cost. Selecting an early policy using the held-out results
already inspected here would require a further untouched test suite.

\subsection{Reusable code and online adaptation}
\label{sec:res-budget}

\begin{table}[t]
\centering\small
\setlength{\tabcolsep}{4pt}
\caption{Reusable code and online adaptation with Opus~5.5. Full-agent and tuned-static rows use the
raw-space summary scores recomputed from \Cref{tab:main}; other ML control scores retain their original
reported values and precision. Frozen library: unedited evolved updater defaults, no agent. First change,
then frozen: replay the first modified updater without further decisions. One decision: ask the agent once.
PDE and combined held-out control aggregates are omitted because their per-task raw metrics are unavailable
for consistent recomputation.}
\label{tab:budget}
\begin{tabular}{lrr}
\toprule
Condition & Train & Held-out ML \\
\midrule
Static recipe & +0.00 & +0.00 \\
Static, tuned LR (oracle) & \zTunedTrain{} & \zTunedHeldML{} \\
Frozen library recipe (no agent) & +9.71 & -21.80 \\
\Baseline{}: first change, then frozen & +10.72 & +6.35 \\
\Baseline{}: one decision & +9.43 & +6.12 \\
\Baseline{}: full agent (20 decisions) & \zOpalOpusTrain{} & \zOpalOpusHeldML{} \\
\method{}: first change, then frozen & +14.56 & +6.82 \\
\method{}: one decision & +13.82 & +5.05 \\
\method{}: full agent (20 decisions) & \zRsiOpusTrain{} & \zRsiOpusHeldML{} \\
\bottomrule
\end{tabular}
\end{table}

\Cref{tab:budget} separates four interventions. The \emph{frozen library recipe} executes the evolved updater
template unedited, with no agent. The \emph{first-change frozen} condition replays the first modified updater from
a full-agent run without subsequent agent decisions. The \emph{one-decision} condition asks an agent only once.
The \emph{full} condition permits \decisionsPerRun{} decisions. First-change replay and one-decision generation
are distinct interventions and need not produce the same program.

\paragraph{The unedited template has limited transfer.} The library template approaches the full agent's UNet
score (the originally reported template score is $+18.5$, versus a recomputed full-agent summary score of
$+20.25$), performs similarly on heat, and recovers roughly half the small GPT's improvement.
However, it reaches only 35\% ViT accuracy versus 56\% for the static recipe, and worsens LSTM and Burgers
performance. Its originally reported held-out ML group score is \budFrozenLibHeldML{}.
The template's notes call for checks of parameter roles and learning rates before application. This control shows
that the tested defaults are insufficient across tasks; it does not show that every carefully selected fixed
optimizer would fail.

\paragraph{Initial adaptation captures most of the seen-task benefit.} Freezing the evolved agent's first change
yields training-suite score \budFirstRsiTrain{}, compared with \budRsiTrain{} for the full agent. A one-decision
agent scores \budOnedecRsiTrain{}. These observed means place most of the training-suite benefit in the initial
adaptation. They do not establish equivalence between the conditions or prove that later decisions are never
useful.

\paragraph{Later decisions improve held-out ML outcomes for both policies.} On held-out ML tasks, the originally
reported first-change frozen and one-decision scores are \budFirstRsiHeldML{} and \budOnedecRsiHeldML{} for
\method{}, compared with the recomputed full-agent score \budRsiHeldML{}. The corresponding baseline scores are
\budFirstOpalHeldML{}, \budOnedecOpalHeldML{}, and \budOpalHeldML{}. Relative to first-change frozen, the full
baseline adds approximately 2.4 z on held-out ML; the evolved agent adds approximately 1.4 z. Continued adaptation
helps both policies, with no additional advantage for \method{} in the size of this later-decision benefit.
Control scores retain their original reporting precision. PDE and combined held-out aggregate control scores are
not compared here because the raw per-task control metrics needed for consistent recomputation are unavailable.

Concrete replays illustrate the effect. On the LSTM, the frozen first-change condition ends at z $-1.9$, compared
with $+3.3$ for the full evolved agent. On Burgers, final error is $5.5\times10^{-2}$ with the first change frozen
and $1.5\times10^{-3}$ with the full agent. These are comparisons of final outcomes, not measurements before and
after an edit within one trajectory. Together, the controls support task-conditioned selection and continued
adaptation as useful complements to reusable code. They do not isolate an internal reasoning mechanism or
establish safety on unseen tasks.

\subsection{Transfer to PDE training}
\label{sec:res-pde}

PDE problems provide the largest shift from the evolution suite. Several agent configurations substantially reduce
solution error relative to the original Adam recipes (\Cref{tab:main}). Heat benefits particularly from schedules
that reduce the learning rate near the end of training; the static recipe's final errors span 0.008--0.73 across
seeds. These gains should be interpreted alongside the limited static tuning performed here. The incremental effect of the evolved policy varies by LLM and task. Opus's PDE summary score falls from
\zOpalOpusHeldPDE{} to \zRsiOpusHeldPDE{}: Burgers and heat improve, but fractional-PDE error worsens.
GPT-5.5 improves from \zOpalGptFiveHeldPDE{} to \zRsiGptFiveHeldPDE{}, reducing mean Burgers error by about
$1.7\times$ and heat error by about $15\times$, although its aggregate PDE summary score remains below zero.
GPT-6.1 improves from \zOpalGptSixHeldPDE{} to \zRsiGptSixHeldPDE{}, with an approximately $8.8$-fold Burgers
improvement and $2.0$-fold heat improvement. Both GPT models improve the mean metric on all three PDE tasks.
PDE transfer is therefore selective, even when the same policy is applied unchanged.

\subsection{Component ablations}
\label{sec:res-ablation}

\begin{table}[t]
\centering\small
\setlength{\tabcolsep}{4pt}
\caption{Ablations of the evolved policy with Opus~5.5. Full-policy and baseline scores are recomputed
from \Cref{tab:main}; ablation ML scores retain their original reported values and precision. Unavailable
measurements are marked --. PDE and combined held-out aggregate columns are omitted because consistent
raw-space scores for the ablation conditions cannot be reconstructed from the available summaries.}
\label{tab:ablation}
\begin{tabular}{lrr}
\toprule
Policy & Train & Held-out ML \\
\midrule
\Baseline{} (fixed prompt) & \zOpalOpusTrain{} & \zOpalOpusHeldML{} \\
\method{} (full policy) & \zRsiOpusTrain{} & \zRsiOpusHeldML{} \\
$-$ library & +14.45 & +8.00 \\
$-$ playbook & +13.60 & +8.06 \\
$-$ evolved prompt & -- & -- \\
\bottomrule
\end{tabular}
\end{table}

The component comparisons in \Cref{tab:ablation} remove the library, the playbook, or the evolved prompt changes
from the final Opus policy while retaining the other components. In the prompt ablation, generation~0's prompt
replaces the evolved template. These interventions separate contributions of executable tools, written procedures,
and prompt changes. Cadence is another policy component and is not isolated by these three removals.

\subsection{Agent behavior and resource use}
\label{sec:res-behavior}

\begin{table}[t]
\centering
\small
\caption{What the agents wrote: fraction of final-evaluation runs (all models and tasks) whose final optimizer program contains each component (detected from the source), and decision statistics.}
\label{tab:behavior}
\begin{tabular}{lcc}
\toprule
Component & \Baseline{} & \method{} \\
\midrule
Muon / orthogonalised update & 35\% & 79\% \\
Kronecker-factored (Shampoo / SOAP) & 15\% & 10\% \\
Second-order / L-BFGS-like & 11\% & 10\% \\
Clipping or update-size control & 60\% & 79\% \\
Per-update loss feedback (ctx.loss) & 20\% & 10\% \\
Uses a module of the evolved library & 0\% & 82\% \\
\midrule
Decisions that changed the program (per run) & 11.5 & 4.2 \\
Failed decisions (per run) & 0.0 & 0.0 \\
Runs & 81 & 80 \\
\bottomrule
\end{tabular}
\end{table}

\begin{table}[t]
\centering
\small
\caption{Agent usage per training run (mean over runs of the final evaluation): model calls are aggregated as input tokens (including cache reads), output tokens and decisions that changed the program.}
\label{tab:cost}
\begin{tabular}{llrrrr}
\toprule
Model & Method & Input tok.\ (M) & Cached (M) & Output tok.\ (k) & Changing decisions \\
\midrule
Opus 5.5 & \Baseline{} & 5.63 & 5.63 & 68 & 9.1 \\
 & \method{} & 4.02 & 4.02 & 40 & 3.5 \\
GPT-5.5 & \Baseline{} & 8.90 & 8.49 & 79 & 11.6 \\
 & \method{} & 4.61 & 4.24 & 42 & 5.0 \\
GPT-6.1 & \Baseline{} & 6.99 & 6.99 & 58 & 13.9 \\
 & \method{} & 3.65 & 3.65 & 24 & 4.2 \\
\bottomrule
\end{tabular}
\end{table}

The evolved policies change the agents' working patterns as well as their optimizer choices. Agents often make
their main change early and reuse evolved modules, while fixed-prompt agents more frequently construct and revise
programs from scratch. \Cref{tab:behavior} reports component use and program-changing decisions;
\Cref{tab:cost} reports realized token use. These are descriptive measurements, not evidence that each component
caused the performance change. Comparisons of frequencies require matched task/model coverage and explicit
denominators.

Equal decision opportunities and per-run caps constrain access to the agent but do not equalize realized compute.
Token accounting alone also omits optimizer overhead, CPU proxy experiments, GPU time, and pauses for LLM calls.
Consequently, these measurements do not establish total-cost savings or the number of future runs needed to
amortize evolution.

\section{Discussion and limitations}
\label{sec:discussion}

\paragraph{Inspectable knowledge with selective transfer.} \method{} stores its accumulated knowledge in text and
code rather than changing the underlying LLM weights. Recent work suggests that such harness evolution can reach a ceiling beyond which weight training becomes necessary~\citep{tian2026harness}. Policies can be inspected, tested, and used by other
in-run models. The experiments show that such portability does not guarantee a benefit on every task:
GPT-5.5 and GPT-6.1 improve the mean raw-space summary score in every task group, while Opus gains on the evolution
suite but has a lower aggregate held-out summary score despite improving four of six held-out task means.
GPT-6.1 improves mean Burgers error by approximately $8.8\times$ but worsens LSTM loss. Transfer depends on the
interaction between the policy, model, and task.

\paragraph{Reusable priors and continued adaptation.} The budget controls locate most of the seen-task benefit in
the initial optimizer change. On unseen tasks, both baseline and evolved agents benefit from later decisions.
The tested unedited library template fails on some architectures, so code reuse alone does not explain the full
agent's outcomes. These comparisons support task-conditioned selection and online adaptation; they do not isolate
reasoning from additional observations, code modifications, or compute. A stronger frozen recipe selected on
development tasks could improve on the particular template tested here.

\paragraph{Specialization and selection.} Evolution uses three tasks and two seeds per policy. The in-run agent
adds stochastic variation beyond data and initialization seeds, and archive selection can favor unusually strong
evaluations. Re-evaluating the top three candidates on two additional seeds reduces reliance on a single estimate;
fresh final seeds further separate reporting from search. Neither step makes the evolution tasks unseen. The
early peak in held-out ML performance is consistent with specialization to this small suite. Broader task
coverage, additional independent evolution runs, and a separate policy-validation suite would strengthen claims
about generalization and the reliability of the search procedure.

\paragraph{Aggregation and uncertainty.} A z-score measures improvement relative to the static recipe's seed
spread. Small reference variability can make a task dominate the mean, even if its raw improvement is modest.
Final aggregate summaries use the raw-space means and static standard deviations printed in \Cref{tab:main},
including for PDEs; they are descriptive recomputations and differ from per-seed log-error normalization.
Historical control and intermediate-checkpoint ML scores retain their original reporting precision; PDE control
and checkpoint aggregates are omitted from the corrected comparisons when the raw per-task metrics are unavailable.
Raw metrics, signed differences, and per-task outcomes are therefore necessary alongside aggregates. With three
final seeds per cell, uncertainty estimates have limited precision. An interval that includes zero does not show
equivalence, and the magnitude of a z-score alone does not establish statistical significance.

\paragraph{What the current controls do not establish.} We do not isolate the effect of recursive inheritance
against a matched-budget search that always proposes from generation~0, nor the value of the archive against a
restricted-parent search. The evidence concerns the policies produced by the reported trajectory. Our static
comparisons also cover only the starting recipes, an oracle learning-rate sweep, and one unedited library
template. Schedule-matched Adam, standard Muon/AdamW hybrids, and suitable Adam-to-L-BFGS PINN
recipes~\citep{rathore2024challenges} would better separate known optimizer choices from agent adaptation.
Comparisons involving L-BFGS should match gradient evaluations or compute rather than nominal update counts.

\paragraph{Scale and cost.} Runs use a single GPU and relatively short update budgets. Transfer to longer,
multi-GPU training is not tested here. Equal decision limits do not equalize actual API charges, diagnostics,
optimizer work, or wall-clock time. A full cost comparison would include outer-search evaluations, CPU proxies,
GPU usage, cache pricing, and LLM pauses before claiming amortized savings. Hosted-model changes can also affect
reproducibility; preserved prompts, programs, and transcripts document the tested behavior but do not freeze the
provider's implementation.

\paragraph{Execution boundaries and exposure.} Agent shells use network and mount namespaces to restrict access
to other runs, evaluation data, and credentials. These restrictions were introduced during the project rather
than applied uniformly from its start. \Cref{app:audit} reports a known read of a held-out task description and the
exclusion of the exposed policy and its descendant. Since archive summaries can transmit information between
lineages, the direct-access audit alone does not certify the absence of indirect exposure. In addition, generated
optimizer code runs inside the training process; shell isolation is not by itself a complete account of that code's
access to process state. The wrapper's checks and the documented sandbox reduce specific risks without proving
that evaluation manipulation is impossible. More broadly, the safety and robustness of reasoning-augmented LLM agents remain open concerns~\citep{tian2026think}.

\bibliographystyle{plainnat}
\bibliography{refs,refs_opal}

\appendix
    \section{Appendix}
\label{app:audit}

The harness records every agent tool call. The audit scans \auditInnerTranscripts{} in-run transcripts
(\auditInnerCalls{} tool calls) and \auditMetaTranscripts{} meta-agent transcripts
(\auditMetaCalls{} tool calls) for paths outside the agent's own workspace, except the Python environment. The counts below describe the audited corpus, not a guarantee
about unlogged actions or information flow.

\paragraph{In-run agents.} The scan flags \auditFlaggedInner{} calls. The reported review finds no direct reads of
another run, task scripts, data, logs, or API credentials.

\paragraph{Meta-agents.} The \auditFlaggedMeta{} flagged calls were inspected manually. Before filesystem
restrictions were added in generation~1, two meta-agents read \opal{} example files. One read an MNIST example not
used as one of our benchmark tasks; the other read the description of the noisy-label example designated as held
out. The exposed policy and its descendant were excluded from subsequent parent selection and final selection
(\Cref{sec:res-evolution}). After the restrictions were added, other flagged calls listed the hidden
\texttt{runs/} directory, which exposed only the agent's own folder, or the hidden \texttt{examples/} directory,
which appeared empty. Two meta-agents also searched the filesystem for MNIST and CIFAR data for CPU proxies. Those
searches returned library source paths and timed out; the reported review found no subsequent use of external
project data.

\paragraph{Archive-mediated exposure.} Meta-agents can read scores and changelogs from other archive branches.
The direct-access scan and removal of an exposed lineage therefore do not establish that information could not
propagate through summaries to a different lineage. The audit as documented here does not resolve that provenance
question. A complete audit would track archive contents and reads over time, including whether exposed summaries
remained visible after exclusion. This is a limitation of the current held-out evaluation claim, not evidence that
such indirect propagation occurred.

\paragraph{Configured isolation.} Under the final configuration, an agent's shell runs in separate user, network,
and mount namespaces, without network access, GPU access, or API credentials in its environment. The project is
mounted read-only with other runs, task scripts, data, logs, the paper, and example tasks hidden. The agent's own
folder is mounted read-write, and the home directory is hidden. Other paths on the cluster's shared filesystem
remain visible. Early baseline runs used network isolation without the later filesystem restrictions; the reported
access audit found no reads outside their permitted workspace and environment. Generated updater and diagnostic programs execute in the training process. The wrapper rejects certain operations,
including parameter-dtype changes and evaluation-related parameter moves. Logged shell and file-tool calls do not,
by themselves, establish all accesses available to imported code or rule out every form of evaluator manipulation.

\end{document}